\documentclass[conference]{IEEEtran}
\IEEEoverridecommandlockouts

\usepackage{cite}
\usepackage{amsmath,amssymb,amsfonts}
\usepackage{algorithmic}
\usepackage{graphicx}
\usepackage{textcomp}
\usepackage[table]{xcolor}
\usepackage{array}
\usepackage{booktabs}
\usepackage{multirow}
\usepackage{adjustbox}
\usepackage{url}
\newcommand{\sacqcodeurl}{https://github.com/PLACEHOLDER/SACQ}
\newcommand{\sacqcodefootnote}{\protect\url{\sacqcodeurl}}

\newcommand{\corruptionheader}[3]{%
  \rowcolor{black!8}%
  Dataset & & \multicolumn{6}{c|}{#1} & \multicolumn{6}{c|}{#2} & \multicolumn{6}{c}{#3} \\
  \rowcolor{white}%
  Model & & \multicolumn{2}{c|}{PatchSACQ} & \multicolumn{2}{c|}{PatchLGA} & \multicolumn{2}{c|}{PatchTST} &
            \multicolumn{2}{c|}{PatchSACQ} & \multicolumn{2}{c|}{PatchLGA} & \multicolumn{2}{c|}{PatchTST} &
            \multicolumn{2}{c|}{PatchSACQ} & \multicolumn{2}{c|}{PatchLGA} & \multicolumn{2}{c}{PatchTST} \\
  \cmidrule(lr){3-4}\cmidrule(lr){5-6}\cmidrule(lr){7-8}%
  \cmidrule(lr){9-10}\cmidrule(lr){11-12}\cmidrule(lr){13-14}%
  \cmidrule(lr){15-16}\cmidrule(lr){17-18}\cmidrule(lr){19-20}%
  Metric & & MSE & MAE & MSE & MAE & MSE & MAE &
            MSE & MAE & MSE & MAE & MSE & MAE &
            MSE & MAE & MSE & MAE & MSE & MAE \\
  \cmidrule(lr){3-4}\cmidrule(lr){5-6}\cmidrule(lr){7-8}%
  \cmidrule(lr){9-10}\cmidrule(lr){11-12}\cmidrule(lr){13-14}%
  \cmidrule(lr){15-16}\cmidrule(lr){17-18}\cmidrule(lr){19-20}%
  \midrule
}
\def\BibTeX{{\rm B\kern-.05em{\sc i\kern-.025em b}\kern-.08em
    T\kern-.1667em\lower.7ex\hbox{E}\kern-.125emX}}
\begin{document}

\title{SACQ: Structured Decoding with Memory-Conditioned Refinement for Long-Horizon Forecasting}

\author{\IEEEauthorblockN{Guo Cheng$^{1}$, Zhengzhuo Xu$^{1}$, Chenchen Jing$^{2}$, Jingyi Hou$^{1,3,*}$}
\IEEEauthorblockA{$^{1}$\textit{University of Science and Technology Beijing, China}\\
$^{2}$\textit{Zhejiang University of Technology, Hangzhou, China}\\
$^{3}$\textit{Institution of Artificial Intelligence, University of Science and Technology Beijing, China}\\
Email: \{u202343375, u202142443\}@xs.ustb.edu.cn, jingchenchen96@163.com, houjingyi@ustb.edu.cn}
\thanks{$^{*}$Corresponding author.}
\thanks{Code is available at \sacqcodefootnote.}
}

\maketitle

\begin{abstract}
Long-term time series forecasting (LTSF) models predominantly employ patch-based encoders terminated by a flatten readout head that maps the entire encoded historical memory to all future steps through a single shared projection. This implicit coupling of future positions obscures position-specific historical-to-future alignment and amplifies sensitivity to corrupted inputs and extreme supervision noise. We present SACQ, a plug-in structured prediction head that replaces flatten readout while keeping the encoder unchanged. SACQ adopts a two-stage decoding pipeline: it first establishes a coarse patch-grid forecast scaffold, then refines each future position through cross-attention over historical memory and merges the attention-derived correction with the coarse scaffold via a learned per-patch gate. To stabilize optimization under long horizons and noisy labels, we further propose a batch-adaptive scaled log-cosh loss that automatically calibrates robustness to the current residual scale, suppressing outlier gradients while preserving MSE-like sensitivity for typical errors. 
SACQ attains top-tier test MSE/MAE across PatchTST, DLinear, and patch-Mamba backbones with only modest incremental overhead in parameters and latency.
Under inference-time input corruption and training-set label-noise stress tests, SACQ substantially outperforms flatten readouts, with ablation studies validating each architectural component.
\end{abstract}


\sloppy

\section{Introduction}

Long-term time series forecasting (LTSF) underpins decision making and planning in domains where forward-looking information is essential, including energy load dispatch \cite{syed2021load}, traffic flow management \cite{li2018dcrnn}, and industrial equipment telemetry. The central challenge is to extrapolate reliably over a distant future from a finite history of observations \cite{wen2023transformerssurvey}. Recent progress has strengthened encoder-side temporal modeling through strong baselines such as PatchTST, Autoformer, and DLinear, which exploit patch tokenization \cite{nie2023patchtst}, frequency-domain decomposition \cite{wu2021autoformer}, and efficient linear baselines \cite{zeng2023transformers}. Yet an often overlooked bottleneck remains at the readout stage. Most models still terminate with a lightweight flatten readout that vectorizes encoded historical patch memory and maps it to the entire forecast horizon through one shared dense projection \cite{nie2023patchtst,zeng2023transformers}. This design is parameter efficient, but it introduces two deeply coupled limitations that become especially harmful under long horizons and imperfect data.

First, \emph{representation coupling}. Because all future positions share the same readout parameters, historical-to-future correspondence is compressed into static weights rather than realized through explicit, position-specific retrieval and alignment, unlike decoder-style forecasters that treat multi-step prediction as structured readout \cite{oreshkin2020nbeats,lim2021tft,wu2021autoformer}. When the input history contains abnormal or corrupted patches, flatten readout cannot apply differentiated readout logic at different future positions, and models therefore degrade quickly under inference-time input perturbations, a pattern also reported under standardized corruption benchmarks \cite{kim2026local}. Second, \emph{optimization coupling}. Under standard mean-squared error (MSE), gradient magnitude scales with residual size, so a small number of extreme supervision steps can dominate updates within a minibatch. The global projection in flatten readout further amplifies this effect, leading to marked test-error increases when training labels contain spikes or tail contamination, with the gap widening as the forecast horizon grows.

\begin{figure*}[t]
  \centering
  \includegraphics[width=0.87\textwidth]{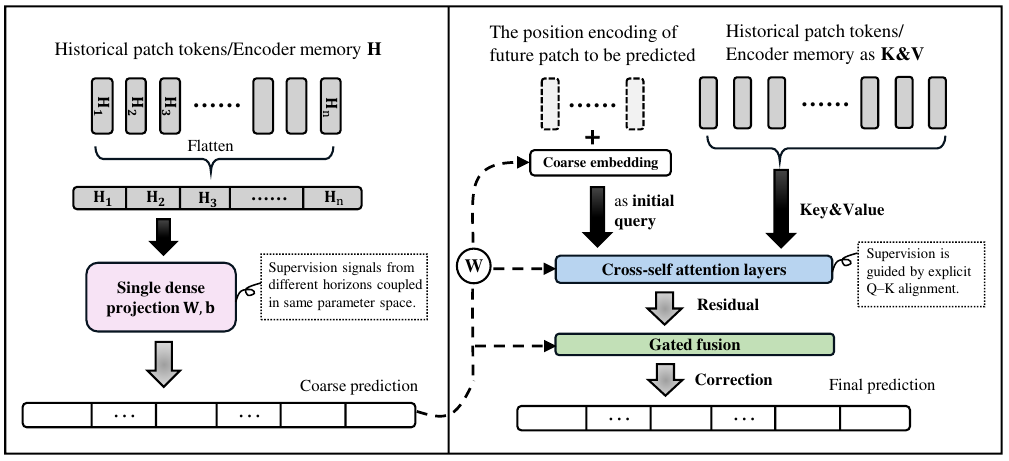}
  \caption{\textbf{Flatten readout vs.\ SACQ decoding.} \textit{Left:} flatten maps encoded historical memory to all future steps through one dense projection, sharing readout parameters across the horizon. \textit{Right:} SACQ initializes coarse future-patch predictions, refines them with cross-self attention over historical memory, and merges a gated residual correction at each future patch. Details are given in Sec.~\ref{sec:sacq_head}.}
  \label{fig:intro_flatten_vs_sacq}
\end{figure*}

To address these limitations systematically, we propose SACQ (Structured Decoding with Memory-Conditioned Refinement), a plug-in structured prediction head that replaces flatten readout without modifying the patch encoder. SACQ recasts long-horizon forecasting as explicit staged decoding. Stage one builds a coarse patch-grid scaffold through a lightweight projection that preserves global structure. Stage two applies cross-attention \cite{vaswani2017attention} so each future position queries historical memory to produce a position-specific correction residual. Stage three merges the attention-derived correction with the coarse scaffold through learned per-patch gating. For training under long horizons and noisy labels, we further adopt batch-adaptive scaled log-cosh loss \cite{saleh2024logcosh}, which sets the width of the robust transition region from the batch median residual and dampens gradients from extreme outliers while retaining MSE-like sensitivity for typical errors, without dataset-specific calibration.

Our contributions are threefold. First, we characterize flatten-readout limitations as representation and optimization coupling and quantify them with inference-time corruption tests, training-set input masking, and label-noise stress experiments. Second, we design SACQ as a structured decoder that integrates coarse initialization, memory-conditioned cross-attention, concat--linear fusion, gated residual readout, and optional causal self-attention, trained by default with batch-adaptive scaled log-cosh. Third, on standard long-horizon benchmarks and readout swaps across PatchTST, DLinear, and patch-Mamba, SACQ achieves top-tier or near top-tier test MSE and MAE under both clean evaluation and robustness settings, with only moderate parameter and latency overhead. Follow-on ablation studies and efficiency analysis further isolate module contributions and deployment cost.

\section{Related Work}
\subsection{Backbone Paradigms for LTSF}
The LTSF literature now develops along multiple backbone families rather than a single dominant paradigm. A major line focuses on Transformer-based designs, which improve long-range forecasting through efficient attention, decomposition, and frequency-domain priors (e.g., LogTrans, LSINet, Pyraformer, Autoformer, TimesNet, and FEDformer) \cite{li2019enhancing,zhang2025lsinet,liu2022pyraformer,wu2021autoformer,wu2023timesnet,zhou2022fedformer}. Building on this trend, subsequent methods place more emphasis on cross-variable dependency modeling, non-stationarity mitigation, and multi-scale temporal structure, as represented by Crossformer, non-stationary Transformers, SCINet, TimeMixer, FILM, and ETSformer \cite{zhang2023crossformer,liu2022nonstationary,liu2022scinet,wang2024timemixer,zhou2022film,woo2023etsformer}. In parallel, complementary perspectives continue to shape the field, including survey-level syntheses and probabilistic forecasting viewpoints \cite{wen2023transformerssurvey,salinas2020deepar}, strong linear/decomposition baselines \cite{zeng2023transformers}, state-space backbones such as Mamba \cite{gu2024mamba}, and recent long-context or efficiency-oriented models such as Timer-XL and TimePro \cite{wang2025timerxl,shi2025timepro}. Taken together, these results suggest that forecasting limitations are unlikely to be tied to any single encoder family, which in turn motivates decoder-side redesign.

\subsection{Prediction Heads as Structured Decoders}
Prediction heads are commonly implemented as lightweight readout modules, such as linear layers, multilayer perceptrons, or one-shot flattening projections. Recent analyses report that simple readouts can remain highly competitive \cite{zeng2023transformers}, suggesting that prediction-side design may be a limiting factor under long horizons. Prediction-side operations can also include normalization and re-scaling around readout (e.g., RevIN) \cite{kim2022revin}, which can be treated as part of the prediction pipeline but do not by themselves define explicit historical-to-future alignment. Beyond one-shot readout, several forecasting Transformers adopt explicit decoder-style designs to support multi-step prediction \cite{wu2021autoformer,zhou2021informer,zhou2022fedformer}. In parallel, strong non-Transformer forecasters such as N-BEATS and TFT highlight the role of decoding inductive bias in multi-horizon prediction \cite{oreshkin2020nbeats,lim2021tft}. Collectively, these lines of work motivate structured prediction heads that make historical-to-future correspondence more explicit than a single dense readout.

\subsection{Robustness Evaluation for LTSF}
Most long-horizon benchmarks still rank models on clean splits \cite{wen2023transformerssurvey}, while distribution-shift methods such as RevIN and non-stationary Transformers \cite{kim2022revin,liu2022nonstationary,zhao2025shifts} target covariate drift rather than localized input defects or label noise. Kim et al.~\cite{kim2026local} recently introduce TSRBench with severity-controlled spike and level-shift corruptions at inference time, showing that flatten readouts can degrade rapidly under realistic perturbations. Robust M-estimators such as log-cosh \cite{saleh2024logcosh} offer a complementary training-side remedy for outlier supervision. Readout-focused stress tests under both regimes nevertheless remain scarce.

\section{Method}
\subsection{Problem Definition}
We address multivariate long-term forecasting. A single example consists of $C$ jointly observed measurement channels sampled on a uniform grid. For each channel, the model conditions on $N$ past observations and predicts the subsequent $T_f$ time steps. When models are trained and evaluated in minibatches, $B$ denotes the batch size. Given these definitions, the input tensor is $\mathbf{X}\in\mathbb{R}^{B\times N\times C}$ and the predicted future tensor is $\hat{\mathbf{Y}}\in\mathbb{R}^{B\times T_f\times C}$, where for each batch element and channel the respective second mode lists time indices in chronological order within the lookback window or the forecast window.

\subsection{Framework Overview}
The forward pipeline follows PatchTST \cite{nie2023patchtst}: reversible instance normalization (RevIN) \cite{kim2022revin}, patch-wise tokenization, backbone encoding of historical patches, decoding with the prediction head, unpatchfying, and inverse RevIN to recover the original scale. RevIN alleviates heterogeneity in magnitude and mitigates moderate non-stationarity across windows. Patch-wise tokenization partitions each channel into contiguous temporal segments that serve as encoder inputs. The backbone maps past patch tokens to a compact memory representation. The prediction head expands this representation into future patch-scale targets, which are subsequently rearranged into per-step forecasts. The backbone is kept fixed while only the flatten readout is replaced with SACQ, which starts from a coarse prediction formed analogously to flatten, maps it to coarse content embeddings, yields a residual through cross--self attention by attending from future queries to encoded memory with keys and values drawn from that memory, and fuses this residual with the coarse prediction through gating (Sec.~\ref{sec:sacq_head}). The overall architecture follows this staged coarse-to-fine decoding workflow.
\begingroup
  \setlength{\intextsep}{5pt plus 2pt minus 2pt}
  \vspace{-2pt}%

\endgroup

\subsection{SACQ Head}
\label{sec:sacq_head}
\subsubsection{Methodology intuition}
SACQ treats long-horizon forecasting as structured decoding at readout: it keeps an ordered patch-token grid and updates future positions with explicit token-level operations instead of one dense projection to the full horizon. Concretely, decoding proceeds in three steps: (i) build a coarse patch-grid scaffold, (ii) apply cross-attention to historical memory for position-specific refinement~\cite{vaswani2017attention}, and (iii) merge coarse and refined signals through a learned gate for residual correction. This makes historical-to-future correspondence an explicit pathway while preserving horizon and patch structure. Only the prediction head is redesigned and the patch encoder stays fixed.

\subsubsection{Design principles}
\textit{P1: two-stage decoding} reduces one-shot regression difficulty by separating patch-grid coarse structure from memory-conditioned refinement. \textit{P2: explicit query--key alignment} restores historical-to-future correspondence as computation, not hidden parameter patterns. \textit{P3: joint re-encoding of coarse priors and cross-attention outputs} combines these signals through feature concatenation followed by linear projection, keeping coarse scaffolding and high-order interactions in one space before self-attention and mitigating brittle propagation when horizons are long. \textit{P4: gated residual readout} fuses the coarse patch forecast and the attention-produced prediction residual under a learned gate, so the final output is a gated mixture of coarse structure and refinement.

\subsubsection{Coarse Structural Initialization} Let $\mathbf{H}\in\mathbb{R}^{N_{\mathrm{p}}\times D}$ denote encoded historical patch memory comprising $N_{\mathrm{p}}$ tokens formed from the length-$N$ input under patch tokenization. Let $N_f$ denote the number of future patch tokens implied by the prediction horizon $T_f$, and let $p$ denote the patch length. Under the same contiguous, non-overlapping patching as the encoder, $N_f=\lceil T_f/p\rceil$. For each batch element and channel, the coarse future grid $\mathbf{F}\in\mathbb{R}^{N_f\times p}$ is obtained by mapping $\operatorname{vec}(\mathbf{H})\in\mathbb{R}^{N_{\mathrm{p}}D}$ through an affine transformation into $\mathbb{R}^{N_f p}$ and reshaping the resulting vector into $N_f$ patches of ambient dimension $p$. Like a flatten head over the full future patch lattice, this parameterization has the same scalar count. Each patch of $\mathbf{F}$ is an unconstrained patch-level coordinate vector until refinement. Query-side content embeddings are produced by applying an affine map independently to each patch of $\mathbf{F}$,
\begin{equation}
  \label{eq:c_from_f}
  \mathbf{C} = \mathbf{F}\mathbf{W}_{c} + \mathbf{b}_{c},
\end{equation}
where $\mathbf{W}_{c}\in\mathbb{R}^{p\times D}$ and $\mathbf{b}_{c}\in\mathbb{R}^{D}$ are shared across all $N_f$ future-patch positions. $\mathbf{C}\in\mathbb{R}^{N_f\times D}$ lifts $\mathbf{F}$ from width $p$ to width $D$. Positional codes for future queries are drawn from an extended table $\mathbf{E}\in\mathbb{R}^{(N_{\mathrm{p}}+N_f)\times D}$ whose index set continues the ordering of encoder patch positions, and the segment associated with future patches is $\mathbf{E}_{N_{\mathrm{p}}:N_{\mathrm{p}}+N_f-1}\in\mathbb{R}^{N_f\times D}$. Decoder queries are initialized by
\begin{equation}
  \label{eq:q0_init}
  \mathbf{Q}^{(0)} = \mathbf{C} + \mathbf{E}_{N_{\mathrm{p}}:N_{\mathrm{p}}+N_f-1},
\end{equation}
which superposes the coarse content embedding with the continuation of the positional encoding beyond the historical patch range.

\subsubsection{Cross-self Coupling} 
\begin{figure*}[!t]
  \centering
  \includegraphics[width=0.87\textwidth]{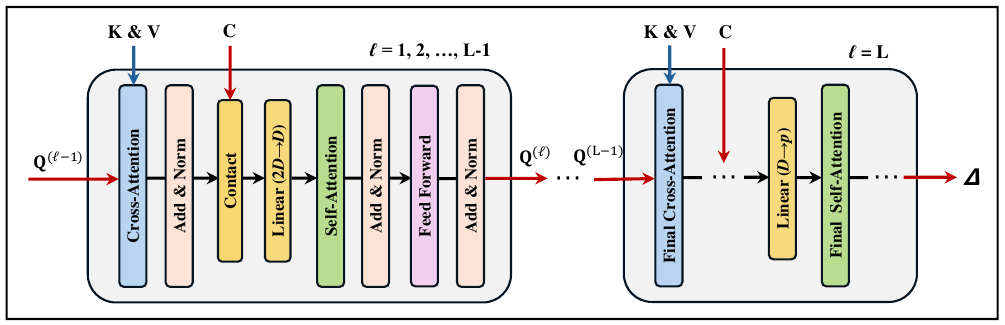}
  \caption{\textbf{Cross-self attention module structure.} \textit{Left:} the first $L{-}1$ layers, each at hidden width $D$ (cross-attention to $\mathbf{H}$, coarse fusion as in Eq.~\eqref{eq:concat_fusion}, self-attention, and FFN). \textit{Right:} the final ($L$-th) layer follows the same high-level pipeline; the ellipsis inside the $L$-th block denotes submodules that are structurally identical to those in the preceding layers. In the last layer only, a linear layer maps from width $D$ to width $p$ \emph{before} self-attention; the self-attention and FFN sublayers then run in the $p$-dimensional space. After the final FFN, the stack outputs $\boldsymbol{\Delta}$ for Eq.~\eqref{eq:residual_readout}.}
  \label{fig:crossself_detail}
\end{figure*}

The cross-self decoder stack is designed to produce a residual tensor $\boldsymbol{\Delta}\in\mathbb{R}^{N_f\times p}$ over the coarse scaffold $\mathbf{F}$. For decoder layer index $\ell=1,\ldots,L$ with $L$ stacked cross-self layers, let cross-attention with keys and values drawn from $\mathbf{H}$ yield the usual sublayer output. Let $\widetilde{\mathbf{Q}}^{(\ell)}\in\mathbb{R}^{N_f\times D}$ denote the post-normalization representation after the residual connection and layer normalization, which serves as the refined future-query representation at layer $\ell$. We couple cross-attention activations with the coarse content embedding $\mathbf{C}$ by first concatenating them along the feature dimension and then applying a learnable linear projection to the concatenated features. Let $\mathbf{U}^{(\ell)}\in\mathbb{R}^{N_f\times D}$ denote the coupled representation at layer $\ell$. Accordingly,
\begin{equation}
  \label{eq:concat_fusion}
  \mathbf{U}^{(\ell)} = \big[\widetilde{\mathbf{Q}}^{(\ell)} \,\|\, \mathbf{C}\big]\,\mathbf{W}^{(\ell)} + \mathbf{b}^{(\ell)},
\end{equation}
where $\mathbf{W}^{(\ell)}$ and $\mathbf{b}^{(\ell)}$ are shared across all $N_f$ future-patch positions. This coupling encourages synchronization between retrieval evidence and coarse structure at each decoder layer, since each layer jointly re-encodes memory-conditioned updates with the shared coarse scaffold. $\mathbf{U}^{(\ell)}$ is then provided to the self-attention and feed-forward stack for further refinement. The cross-self attention module is illustrated in Fig.~\ref{fig:crossself_detail}.

\subsubsection{Gated Residual Readout} After the final decoder layer, $\boldsymbol{\Delta}$ is aligned with $\mathbf{F}$ at both the patch index and the within-patch coordinate. For each future patch index $i\in\{1,\ldots,N_f\}$, let $\widehat{\mathbf{P}}_{i,:}\in\mathbb{R}^{p}$ denote the fused prediction vector for patch $i$. Residual correction is applied as
\begin{equation}
  \label{eq:residual_readout}
  \widehat{\mathbf{P}}_{i,:} = \mathbf{F}_{i,:} + \alpha_i\boldsymbol{\Delta}_{i,:},
\end{equation}
where $\alpha_i\in(0,1)$ is the gate coefficient for the $i$-th future patch. $\alpha_i$ is computed by concatenating the coarse scaffold patch $\mathbf{F}_{i,:}$ and the residual tensor patch $\boldsymbol{\Delta}_{i,:}$, applying a linear map, and then passing the result through a sigmoid nonlinearity: $\alpha_i=\operatorname{Sigmoid}\!\left([\mathbf{F}\,\|\,\boldsymbol{\Delta}]_{i,:}\,\mathbf{w}_{g}+b_{g}\right)$, with $[\mathbf{F}\,\|\,\boldsymbol{\Delta}]\in\mathbb{R}^{N_f\times 2p}$. The parameters $\mathbf{w}_{g}\in\mathbb{R}^{2p}$ and $b_{g}\in\mathbb{R}$ are shared across all future patches. This readout balances structural stability and adaptive correction by retaining the coarse scaffold as a reliable baseline and letting the residual tensor provide patch-specific adjustments through the learned gate.

\subsubsection{SAR: causal self-attention over future patches}
\label{sec:sar}
Structural Alignment with causal Readout (SAR) preserves the SACQ pipeline in Sec.~\ref{sec:sacq_head}---historical memory $\mathbf{H}$, coarse scaffold $\mathbf{F}$ and content embedding $\mathbf{C}$, cross-attention to $\mathbf{H}$, concat--linear fusion with $\mathbf{C}$ in Eq.~\eqref{eq:concat_fusion}, gated residual readout in Eq.~\eqref{eq:residual_readout}, and non-autoregressive unpatchfying to $\hat{\mathbf{Y}}$---and changes only the self-attention sublayers that operate on future-patch queries.

In the default SACQ decoder, future positions $i,j\in\{1,\ldots,N_f\}$ can interact bidirectionally in self-attention: when refining patch $i$, the representation may already aggregate keys and values from later future patches $j>i$.
Those later queries are themselves conditioned on $\mathbf{H}$ through cross-attention, so training can exploit a non-causal ``future-to-past'' path on the query graph that does not match stepwise generation, and long-horizon drift can blur the division between coarse structure in $\mathbf{F}$ and patch-local correction in $\boldsymbol{\Delta}$.
SAR replaces this with a causal (near-to-far) inductive bias while keeping one-shot inference unchanged.

Let $\mathbf{X}^{(\ell)}\in\mathbb{R}^{N_f\times D'}$ denote the future-side input to the self-attention sublayer at decoder layer $\ell$ ($D'=D$ for layers $1,\ldots,L{-}1$; $D'=p$ in the final layer after mapping to patch space, as in Fig.~\ref{fig:crossself_detail}).
With projections $\mathbf{Q},\mathbf{K},\mathbf{V}$ from $\mathbf{X}^{(\ell)}$, standard attention is $\operatorname{Attn}(\mathbf{Q},\mathbf{K},\mathbf{V})=\operatorname{softmax}(\mathbf{Q}\mathbf{K}^\top/\sqrt{D'})\mathbf{V}$.
SAR applies a causal mask $M_{ij}=0$ if $j<i$ and $M_{ij}=-\infty$ if $j\ge i$, so that
\begin{equation}
\label{eq:sar_mask}
A_{ij}=\operatorname{softmax}_j\!\left(\frac{\mathbf{q}_i^\top\mathbf{k}_j}{\sqrt{D'}}+M_{ij}\right),
\end{equation}
where $\mathbf{q}_i^\top$ and $\mathbf{k}_j^\top$ are the $i$-th and $j$-th rows of $\mathbf{Q}$ and $\mathbf{K}$.
Patch $i$ therefore aggregates context only from strictly earlier future patches ($j<i$); neither position $i$ nor any later patch contributes keys or values to the update at $i$, so information flows monotonically along increasing future-patch index.
Eq.~\eqref{eq:sar_mask} is used in all $L{-}1$ width-$D$ self-attention blocks and in the final width-$p$ self-attention block; cross-attention to $\mathbf{H}$ remains fully connected (every future query may attend to all historical tokens), and Eqs.~\eqref{eq:concat_fusion} and~\eqref{eq:residual_readout} are unchanged.

SAR is not stepwise autoregressive decoding at the output level: a single forward pass still produces the full $\hat{\mathbf{Y}}$.
The mask shapes only the internal dependency graph among future queries---encouraging representations to be refined from near to far---without imposing an explicit chain factorization on scalar or patch outputs.
In summary, historical memory $\mathbf{H}$ is visible to all future patches through cross-attention, while future-patch self-attention follows a directed chain $\text{patch }1 \rightarrow \cdots \rightarrow \text{patch }N_f$ before $\boldsymbol{\Delta}$ is fused with $\mathbf{F}$ through $\alpha_i$ in Eq.~\eqref{eq:residual_readout}.

\subsubsection{Output Reconstruction} For each channel, $\widehat{\mathbf{P}}_{i,:}\in\mathbb{R}^{p}$ are stacked over $i$ to form $\widehat{\mathbf{P}}\in\mathbb{R}^{N_f\times p}$. The resulting patch grid is then reshaped into a one-dimensional forecast sequence in chronological order by reading each patch along its within-patch coordinate and concatenating the $N_f$ patches. Because $N_f=\lceil T_f/p\rceil$, this sequence has length $N_f p\ge T_f$, so only the first $T_f$ positions are retained to match the forecasting horizon. Applying the same reconstruction procedure to all channels and minibatch elements gives $\hat{\mathbf{Y}}\in\mathbb{R}^{B\times T_f\times C}$ in RevIN-normalized space~\cite{kim2022revin}. The targets $\mathbf{Y}\in\mathbb{R}^{B\times T_f\times C}$ are defined in the same normalized space, and inverse RevIN is used only when forecasts on the original scale are required.

\subsection{Training objective}
\label{sec:training_objective}
Mean squared error (MSE) is a standard choice for flatten heads, where future coordinates are usually produced by a single direct projection and optimization is often stable in practice. In SACQ, predictions are formed through staged coarse-plus-residual refinement with patch-specific gating, which can make residual magnitudes more uneven across future positions. Since MSE scales gradients linearly with $|e|$ (with $e=\hat{y}-y$), positions with larger residuals can receive disproportionate updates, especially under long horizons, multivariate targets, and unreliable supervision. This behavior does not invalidate MSE, but it can reduce training robustness for the SACQ head.

We therefore adopt a batch-adaptive scaled log-cosh loss. Standard log-cosh, $\rho(e)=\log\cosh(e)$, is a smooth robust M-estimator---locally quadratic near the origin and asymptotically $L_1$-like in the tails---but its quadratic-to-linear transition is fixed and decoupled from the residual scale of each dataset or training stage~\cite{saleh2024logcosh}. To align robustness with the current data scale, we rescale residuals before applying log-cosh:
\begin{equation}
  \label{eq:logcosh_scaled}
  \ell(e;\beta)=\beta^{2}\,\log\cosh\!\left(\frac{e}{\beta}\right),
\end{equation}
where $\beta>0$ sets the width of the robust transition: errors much smaller than $\beta$ are treated like MSE, whereas errors much larger than $\beta$ receive damped, approximately linear penalties. The $\beta^{2}$ prefactor restores local curvature so that, for $|e|\ll\beta$, $\ell(e;\beta)\approx \tfrac{1}{2}e^{2}$, matching MSE sensitivity in the typical-error regime while still suppressing gradients from mislabeled or extreme target steps.

Rather than tuning $\beta$ per dataset, we estimate it automatically from the current minibatch. Let $e_{b,t,c}=\hat{y}_{b,t,c}-y_{b,t,c}$ denote scalar residuals over batch index $b$, horizon $t$, and channel $c$. We use the batch median
\begin{equation}
  \label{eq:beta_median}
  m=\operatorname{median}_{b,t,c}\big(|e_{b,t,c}|\big),
\end{equation}
which is robust to sporadic outliers in the batch and reflects the typical residual magnitude at the current optimization step. The scale is then set to $\beta=k\,m$ with $k{=}0.5$ by default, so $\beta$ widens early in training when errors are large and narrows later as predictions improve, without dataset-specific calibration. Gradient flow through $\beta$ is stopped ($\beta$ treated as a constant during backpropagation), so the scale only reshapes the loss rather than introducing an auxiliary minimization path. Test metrics remain standard MSE and MAE; only the training objective changes.

The implemented training loss averages Eq.~\eqref{eq:logcosh_scaled} over all forecast elements,
\begin{equation}
  \label{eq:pred_loss_stable}
  \mathcal{L}_{\text{pred}}
  = \frac{1}{B\,T_f\,C}\sum_{b=1}^{B}\sum_{t=1}^{T_f}\sum_{c=1}^{C}
  \ell\!\big(e_{b,t,c};\beta\big),
\end{equation}
with $\beta$ from Eq.~\eqref{eq:beta_median}. For numerical stability we evaluate each $\log\cosh(e/\beta)$ in absolute-residual form, $|e/\beta|+\log(1+e^{-2|e/\beta|})$, omitting the additive constant $-\log 2$ that does not affect gradients. The scaled log-cosh objective (Eqs.~\eqref{eq:logcosh_scaled}--\eqref{eq:pred_loss_stable}) is analyzed in Sec.~\ref{sec:loss_ablation_ett_weather}; component ablations appear in Sec.~\ref{sec:ablation_components}.


\section{Experiments}
\label{sec:experiments}
We evaluate SACQ under a unified protocol organized in four layers: (\textit{i})~\textbf{standard performance} on clean benchmarks and across backbones (Sec.~\ref{sec:main_results}); (\textit{ii})~\textbf{internal mechanisms} via component, objective, and hyperparameter ablations (Sec.~\ref{sec:ablation_study}); (\textit{iii})~\textbf{robustness under corrupted inputs and labels} (Sec.~\ref{sec:robustness_eval}); and (\textit{iv})~\textbf{context sensitivity and deployment cost} (Secs.~\ref{sec:lookback_sensitivity}--\ref{sec:efficiency}). Unless stated otherwise, SACQ uses the PatchTST backbone with scaled log-cosh (Eqs.~\eqref{eq:logcosh_scaled}--\eqref{eq:pred_loss_stable}).

\subsection{Experimental Setup}
\label{sec:exp_setup}

\subsubsection{Datasets.}
We evaluate on ETT (ETTh1, ETTh2, ETTm1, ETTm2), Electricity (ECL), Traffic, and Weather, with train/validation/test splits following the usual long-horizon protocol~\cite{wu2021autoformer,zeng2023transformers,zhou2021informer,li2018dcrnn}. The ETT subsets contain 7 variates each (ETTh1/ETTh2: 17{,}420 timestamps; ETTm1/ETTm2: 69{,}680). Weather has 21 variates over 52{,}696 steps; ECL has 321 channels over 26{,}304 steps; Traffic has 862 series over 17{,}544 steps. We emphasize relative ranking under identical preprocessing rather than cross-dataset calibration of absolute error.

\subsubsection{Implementation Details.}
All models run on Linux workstations (Intel Xeon Gold 6530; NVIDIA RTX 4090). We optimize with Adam, a step-wise learning-rate schedule (constant for three epochs, then $\times 0.9$ per epoch), full precision without AMP, and early stopping on validation error. Batch size is 128 for ETT/Weather and 8 for Traffic/ECL; learning rate is $5{\times}10^{-4}$ for ETT/Weather and $10^{-3}$ for Traffic/ECL. The SACQ head uses $L{=}2$ cross-attention layers and scaled log-cosh with $k{=}0.5$ in $\beta=k\,m$ (Eq.~\eqref{eq:beta_median}); other settings follow PatchTST~\cite{nie2023patchtst}.

\subsubsection{Metrics and Baselines.}
We report test MSE and MAE on forecast targets under the standard normalized protocol. SACQ is trained with scaled log-cosh unless an ablation states otherwise; published baselines use their reference objectives (typically MSE). Table~\ref{tab:main_results} compares SACQ against LSINet~\cite{zhang2025lsinet}, PatchLGA~\cite{kim2026local}, PatchTST~\cite{nie2023patchtst}, CI-TSMixer, FiLM~\cite{zhou2022film}, DLinear~\cite{zeng2023transformers}, FEDformer~\cite{zhou2022fedformer}, and TimesNet~\cite{wu2023timesnet}. Readout swaps on patch-Linear and patch-Mamba backbones appear in Sec.~\ref{sec:exp_cross_backbone}.

\subsection{Main Results}
\label{sec:main_results}

\subsubsection{Standard benchmark comparison.}
Table~\ref{tab:main_results} compares SACQ against published baselines. The first column lists the dataset and the second gives the prediction horizon~$T_f\in\{96,192,336,720\}$. Each method reports one test MSE and one MAE per $(\text{dataset},T_f)$; lower is better. \textbf{Bold} marks the best value and \underline{underline} the second-best at each $(\text{dataset},T_f)$, with MSE and MAE ranked separately across methods.

\begin{table*}[t]
  \centering
  \scriptsize
  \setlength{\tabcolsep}{2.5pt}
  \renewcommand{\arraystretch}{0.95}
  \caption{Comprehensive results of the multivariate long-term forecasting task. ``$-$'' indicates values not reported in the original literature (LSINet/FiLM/CI-TSMixer on Traffic). \textbf{Bold} and \underline{underline} mark best and second-best per metric at each $(\mathrm{dataset},T_f)$. SACQ uses scaled log-cosh (Eqs.~\eqref{eq:logcosh_scaled}--\eqref{eq:pred_loss_stable}).}
  \label{tab:main_results}
  \begin{adjustbox}{width=\textwidth,center}
  \begin{tabular}{@{}>{\raggedright\arraybackslash}p{4.6em}@{\hspace{-2pt}}r@{\hspace{1pt}}|cc|cc|cc|cc|cc|cc|cc|cc|cc@{}}
    \toprule
     &  & \multicolumn{2}{c|}{PatchSACQ} & \multicolumn{2}{c|}{LSINet} & \multicolumn{2}{c|}{PatchLGA} & \multicolumn{2}{c|}{PatchTST} & \multicolumn{2}{c|}{{\scriptsize CI-TSMixer}} & \multicolumn{2}{c|}{FiLM} & \multicolumn{2}{c|}{DLinear} & \multicolumn{2}{c|}{FEDformer} & \multicolumn{2}{c|}{TimesNet} \\
    \cmidrule(lr){3-4}
    \cmidrule(lr){5-6}
    \cmidrule(lr){7-8}
    \cmidrule(lr){9-10}
    \cmidrule(lr){11-12}
    \cmidrule(lr){13-14}
    \cmidrule(lr){15-16}
    \cmidrule(lr){17-18}
    \cmidrule(lr){19-20}
    Dataset & {\scriptsize$T_f$} & MSE & MAE & MSE & MAE & MSE & MAE & MSE & MAE & MSE & MAE & MSE & MAE & MSE & MAE & MSE & MAE & MSE & MAE \\
    \midrule
\multirow{4}{*}{ETTh1} & 96 & \textbf{0.361} & \textbf{0.390} & \underline{0.366} & \underline{0.391} & 0.370 & 0.401 & 0.370 & 0.400 & 0.373 & 0.398 & 0.371 & 0.394 & 0.375 & 0.399 & 0.376 & 0.415 & 0.384 & 0.402 \\
      & 192 & \textbf{0.398} & \underline{0.416} & \underline{0.400} & \textbf{0.412} & 0.423 & 0.436 & 0.413 & 0.429 & 0.405 & 0.427 & 0.414 & 0.423 & 0.405 & \underline{0.416} & 0.423 & 0.446 & 0.436 & 0.429 \\
      & 336 & \textbf{0.417} & \textbf{0.427} & 0.427 & \underline{0.428} & 0.433 & 0.447 & \underline{0.422} & 0.440 & 0.426 & 0.441 & 0.442 & 0.445 & 0.439 & 0.443 & 0.444 & 0.462 & 0.491 & 0.469 \\
      & 720 & \textbf{0.429} & \textbf{0.452} & \underline{0.441} & \underline{0.457} & 0.470 & 0.480 & 0.447 & 0.468 & \underline{0.441} & 0.460 & 0.465 & 0.472 & 0.472 & 0.490 & 0.469 & 0.492 & 0.521 & 0.500 \\
    \midrule
    \multirow{4}{*}{ETTh2} & 96 & \underline{0.269} & \textbf{0.334} & \textbf{0.267} & \underline{0.337} & 0.274 & 0.335 & 0.274 & \underline{0.337} & 0.278 & 0.344 & 0.284 & 0.348 & 0.289 & 0.353 & 0.332 & 0.374 & 0.340 & 0.374 \\
      & 192 & \underline{0.333} & \underline{0.377} & \textbf{0.324} & \textbf{0.372} & 0.340 & 0.378 & 0.341 & 0.382 & \underline{0.338} & 0.382 & 0.357 & 0.400 & 0.383 & 0.418 & 0.407 & 0.446 & 0.402 & 0.414 \\
      & 336 & \textbf{0.324} & \textbf{0.380} & 0.345 & 0.394 & 0.329 & 0.381 & \underline{0.329} & \underline{0.384} & 0.356 & 0.405 & 0.377 & 0.417 & 0.448 & 0.465 & 0.400 & 0.447 & 0.452 & 0.452 \\
      & 720 & \textbf{0.376} & \textbf{0.419} & 0.382 & 0.430 & 0.392 & 0.432 & \underline{0.379} & \underline{0.422} & 0.391 & 0.433 & 0.439 & 0.456 & 0.605 & 0.551 & 0.412 & 0.469 & 0.462 & 0.468 \\
    \midrule
    \multirow{4}{*}{ETTm1} & 96 & \textbf{0.286} & \textbf{0.333} & \underline{0.293} & \underline{0.341} & 0.288 & 0.344 & \underline{0.293} & 0.346 & \underline{0.293} & 0.343 & 0.302 & 0.349 & 0.299 & 0.343 & 0.326 & 0.390 & 0.338 & 0.375 \\
      & 192 & \textbf{0.328} & \textbf{0.361} & \underline{0.329} & \underline{0.363} & 0.333 & 0.370 & 0.333 & 0.370 & 0.331 & 0.365 & 0.338 & 0.373 & 0.335 & 0.365 & 0.365 & 0.415 & 0.374 & 0.387 \\
      & 336 & \textbf{0.356} & \textbf{0.379} & \underline{0.357} & \underline{0.384} & 0.361 & 0.386 & 0.369 & 0.392 & 0.363 & \underline{0.384} & 0.365 & 0.385 & 0.369 & 0.386 & 0.392 & 0.425 & 0.410 & 0.411 \\
      & 720 & \underline{0.413} & \underline{0.412} & \textbf{0.396} & \textbf{0.409} & 0.420 & 0.416 & 0.416 & 0.420 & \underline{0.405} & 0.419 & 0.420 & 0.420 & 0.425 & 0.421 & 0.446 & 0.458 & 0.478 & 0.450 \\
    \midrule
    \multirow{4}{*}{ETTm2} & 96 & \textbf{0.159} & \textbf{0.244} & \underline{0.161} & \underline{0.256} & 0.168 & 0.257 & 0.166 & \underline{0.256} & 0.166 & 0.258 & 0.165 & \underline{0.256} & 0.167 & 0.260 & 0.180 & 0.271 & 0.187 & 0.267 \\
      & 192 & \underline{0.216} & \textbf{0.283} & \textbf{0.213} & \underline{0.295} & 0.225 & 0.295 & 0.223 & 0.296 & 0.222 & 0.296 & 0.222 & 0.296 & 0.224 & 0.303 & 0.252 & 0.318 & 0.249 & 0.309 \\
      & 336 & \underline{0.271} & \textbf{0.320} & \textbf{0.257} & \underline{0.327} & 0.277 & 0.329 & 0.274 & 0.329 & 0.274 & 0.328 & 0.277 & 0.333 & 0.281 & 0.342 & 0.324 & 0.364 & 0.321 & 0.351 \\
      & 720 & \underline{0.358} & \underline{0.375} & \textbf{0.328} & \textbf{0.372} & 0.359 & 0.381 & 0.362 & 0.385 & \underline{0.340} & 0.380 & 0.371 & 0.389 & 0.397 & 0.421 & 0.410 & 0.420 & 0.408 & 0.403 \\
    \midrule
    \multirow{4}{*}{Weather} & 96 & \textbf{0.146} & \textbf{0.185} & \underline{0.147} & 0.199 & \underline{0.147} & 0.198 & 0.149 & 0.198 & \underline{0.147} & \underline{0.197} & 0.199 & 0.262 & 0.176 & 0.237 & 0.238 & 0.314 & 0.172 & 0.220 \\
      & 192 & \textbf{0.192} & \textbf{0.230} & \textbf{0.192} & 0.243 & 0.195 & 0.242 & \underline{0.194} & 0.241 & \textbf{0.192} & \underline{0.240} & 0.228 & 0.288 & 0.220 & 0.282 & 0.275 & 0.329 & 0.219 & 0.261 \\
      & 336 & \underline{0.243} & \textbf{0.271} & \textbf{0.238} & 0.281 & 0.246 & 0.283 & 0.245 & 0.282 & \underline{0.243} & \underline{0.279} & 0.267 & 0.323 & 0.265 & 0.319 & 0.339 & 0.377 & 0.280 & 0.306 \\
      & 720 & \underline{0.310} & \textbf{0.326} & \textbf{0.304} & \underline{0.332} & 0.317 & 0.337 & 0.314 & 0.334 & 0.310 & 0.333 & 0.319 & 0.361 & 0.323 & 0.362 & 0.389 & 0.409 & 0.365 & 0.359 \\
    \midrule
    \multirow{4}{*}{ECL} & 96 & \textbf{0.128} & \textbf{0.218} & \underline{0.129} & 0.224 & 0.130 & 0.225 & \underline{0.129} & \underline{0.222} & 0.131 & 0.226 & 0.154 & 0.267 & 0.140 & 0.237 & 0.186 & 0.302 & 0.168 & 0.272 \\
      & 192 &  \textbf{0.144} & \textbf{0.234} & \textbf{0.144} & \underline{0.240} & 0.148 & 0.242 & \underline{0.147} & \underline{0.240} & \underline{0.147} & 0.244 & 0.164 & 0.258 & 0.153 & 0.249 & 0.197 & 0.311 & 0.184 & 0.322 \\
      & 336 & \underline{0.162} & \textbf{0.251} & \textbf{0.158} & \underline{0.256} & 0.166 & 0.259 & 0.163 & 0.259 & 0.163 & 0.261 & 0.188 & 0.283 & 0.169 & 0.267 & 0.213 & 0.328 & 0.198 & 0.300 \\
      & 720 & {0.198} & \textbf{0.282} & \textbf{0.192} & \underline{0.287} & 0.200 & 0.289 & \underline{0.197} & 0.290 & 0.200 & 0.294 & 0.236 & 0.332 & 0.203 & 0.301 & 0.233 & 0.344 & 0.220 & 0.320 \\
    \midrule
    \multirow{4}{*}{Traffic} & 96 & \textbf{0.356} & \textbf{0.245} & $-$ & $-$ & $-$ & $-$ & \underline{0.360} & \underline{0.249} & 0.370 & 0.258 & 0.416 & 0.294 & 0.410 & 0.282 & 0.576 & 0.359 & 0.593 & 0.321 \\
      & 192 & \textbf{0.374} & \textbf{0.251} & $-$ & $-$ & $-$ & $-$ & \underline{0.379} & \underline{0.256} & 0.390 & 0.268 & 0.408 & 0.288 & 0.423 & 0.287 & 0.610 & 0.380 & 0.617 & 0.336 \\
      & 336 & \textbf{0.387} & \textbf{0.255} & $-$ & $-$ & $-$ & $-$ & \underline{0.392} & \underline{0.264} & 0.404 & 0.276 & 0.425 & 0.298 & 0.436 & 0.296 & 0.608 & 0.375 & 0.629 & 0.336 \\
      & 720 & \textbf{0.427} & \textbf{0.276} & $-$ & $-$ & $-$ & $-$ & \underline{0.432} & \underline{0.286} & 0.443 & 0.297 & 0.520 & 0.353 & 0.466 & 0.315 & 0.621 & 0.375 & 0.640 & 0.350 \\
    \midrule
    \# best &  & \multicolumn{2}{c|}{18/24} & \multicolumn{2}{c|}{12/4} & \multicolumn{2}{c|}{0/0} & \multicolumn{2}{c|}{0/0} & \multicolumn{2}{c|}{1/0} & \multicolumn{2}{c|}{0/0} & \multicolumn{2}{c|}{0/0} & \multicolumn{2}{c|}{0/0} & \multicolumn{2}{c|}{0/0} \\
    Avg.\ rank &  & \multicolumn{2}{c|}{1.55/1.16} & \multicolumn{2}{c|}{1.94/2.46} & \multicolumn{2}{c|}{4.71/4.42} & \multicolumn{2}{c|}{3.61/3.88} & \multicolumn{2}{c|}{3.46/3.82} & \multicolumn{2}{c|}{5.98/5.95} & \multicolumn{2}{c|}{6.29/6.11} & \multicolumn{2}{c|}{7.91/8.45} & \multicolumn{2}{c|}{8.07/7.32} \\
    \bottomrule
  \end{tabular}
  \end{adjustbox}
\end{table*}

Table~\ref{tab:main_results} shows that SACQ is consistently in the top tier across all benchmarks, with most cells achieving either the best or second-best MSE/MAE (18/24 best counts; average rank 1.55/1.16 on MSE/MAE). On Traffic and Electricity, SACQ keeps clear gains on MSE and remains competitive on MAE across long horizons, where the output space is high-dimensional. On Weather, the ranking between SACQ and LSINet is close and occasionally alternates by metric. Within ETT, the improvements are dataset-dependent: ETTh2 and ETTm2 show clearer advantages, while ETTh1 and ETTm1 include near-tie regions where absolute differences are small. Forecasting error increases with $T_f$ across methods, as expected; SACQ's benefit is not confined to short-range prediction---on difficult long-horizon settings such as Traffic/Electricity at $T_f=720$, SACQ still keeps favorable ranks. Among non-SACQ entries, PatchTST remains a consistently strong patchwise Transformer baseline, while LSINet is frequently the closest competitor on Weather and parts of ETT.

\subsubsection{Generalization across backbones.}
\label{sec:exp_cross_backbone}
To verify that the gains in Table~\ref{tab:main_results} are not specific to PatchTST, we swap only the readout under patch-Linear (\textbf{L}) and patch-Mamba (\textbf{M})~\cite{gu2024mamba} encoders with matched ETT splits, horizons $T_f\in\{96,192,336,720\}$, and training budgets. Table~\ref{tab:readout_swap} reports $T_f\in\{96,720\}$; keys \textbf{L}/\textbf{M} = backbone, \textbf{F}/\textbf{S} = flatten (MSE) vs.\ SACQ (scaled log-cosh).

\begin{table}[!t]
  \centering
  \caption{Flatten vs.\ SACQ on ETT ($T_f\in\{96,720\}$) under patch-Linear and patch-Mamba backbones. \textbf{Bold} marks the better MSE/MAE in each backbone pair at each $(\mathrm{dataset},T_f)$. Only the readout changes; budgets match within each backbone.}
  \label{tab:readout_swap}
  \begingroup
  \fontsize{6}{7}\selectfont
  \setlength{\tabcolsep}{1.8pt}
  \renewcommand{\arraystretch}{0.92}
  \begin{adjustbox}{width=0.95\columnwidth,center}
  \begin{tabular}{@{}>{\raggedright\arraybackslash}p{3.8em}@{\hspace{-2pt}}r@{\hspace{1pt}}|cc|cc||cc|cc@{}}
    \toprule
     &  & \multicolumn{2}{c|}{L+F} & \multicolumn{2}{c||}{L+S} & \multicolumn{2}{c|}{M+F} & \multicolumn{2}{c}{M+S} \\
    \cmidrule(lr){3-4}
    \cmidrule(lr){5-6}
    \cmidrule(lr){7-8}
    \cmidrule(lr){9-10}
    Dataset & $T_f$ & MSE & MAE & MSE & MAE & MSE & MAE & MSE & MAE \\
    \midrule
    \multirow{2}{*}{ETTh1} & 96 & 0.383 & 0.408 & \textbf{0.374} & \textbf{0.401} & 0.383 & 0.410 & \textbf{0.377} & \textbf{0.401} \\
     & 720 & 0.487 & 0.496 & \textbf{0.473} & \textbf{0.484} & 0.493 & 0.498 & \textbf{0.423} & \textbf{0.469} \\
    \midrule
    \multirow{2}{*}{ETTh2} & 96 & 0.302 & 0.374 & \textbf{0.295} & \textbf{0.354} & 0.291 & 0.362 & \textbf{0.283} & \textbf{0.355} \\
     & 720 & 0.832 & 0.653 & \textbf{0.819} & \textbf{0.630} & 0.427 & 0.450 & \textbf{0.400} & \textbf{0.444} \\
    \midrule
    \multirow{2}{*}{ETTm1} & 96 & 0.310 & 0.356 & \textbf{0.301} & \textbf{0.345} & 0.297 & 0.346 & \textbf{0.288} & \textbf{0.338} \\
     & 720 & 0.437 & 0.430 & \textbf{0.423} & \textbf{0.421} & 0.433 & 0.422 & \textbf{0.412} & \textbf{0.416} \\
    \midrule
    \multirow{2}{*}{ETTm2} & 96 & 0.193 & 0.288 & \textbf{0.187} & \textbf{0.279} & 0.174 & 0.262 & \textbf{0.167} & \textbf{0.257} \\
     & 720 & 0.582 & 0.477 & \textbf{0.567} & \textbf{0.468} & 0.383 & 0.401 & \textbf{0.347} & \textbf{0.383} \\
    \bottomrule
  \end{tabular}
  \end{adjustbox}
  \endgroup
\end{table}

Because only the readout swaps within each backbone, gains isolate structured decoding from encoder capacity. SACQ lowers both metrics versus flatten on every reported cell. Under patch-Linear, the benefit is largest on difficult ETTh2 long-horizon settings where flatten errors are highest; under patch-Mamba, SACQ wins uniformly and the gap widens at $T_f{=}720$. Together with Table~\ref{tab:main_results}, these results support SACQ as a plug-in head rather than a PatchTST-specific recipe.

\subsection{Ablation Studies}
\label{sec:ablation_study}

We next isolate design choices under the PatchTST backbone and the protocol in Sec.~\ref{sec:exp_setup}. Unless noted, the full SACQ head and scaled log-cosh (Eqs.~\eqref{eq:logcosh_scaled}--\eqref{eq:pred_loss_stable}) are used; decoder depth and width match the full model.

\subsubsection{Component-wise analysis.}
\label{sec:ablation_components}

Table~\ref{tab:component_ablation} ablates individual SACQ modules on ETTh1, ETTm1, Weather, and ECL at $T_f{=}720$. Each variant disables one mechanism while keeping all other settings fixed.

\begin{table}[!t]
    \centering
    \caption{Component ablation at $T_f{=}720$ (PatchTST + scaled log-cosh). \textbf{Bold}: full SACQ. Keys: $-$CA, w/o cross-attention; $-$G, fixed $\alpha{\equiv}1$; Coarse, coarse init only; $-$Fus, w/o concat--linear fusion (Eq.~\eqref{eq:concat_fusion}).}
    \label{tab:component_ablation}
    \begingroup
    \fontsize{6}{7}\selectfont
    \setlength{\tabcolsep}{1.8pt}
    \renewcommand{\arraystretch}{0.92}
    \begin{adjustbox}{width=0.95\columnwidth,center}
    \begin{tabular}{@{}>{\raggedright\arraybackslash}p{3.8em}@{\hspace{-2pt}}r@{\hspace{1pt}}|cc|cc||cc|cc@{}}
      \toprule
       &  & \multicolumn{2}{c|}{ETTh1} & \multicolumn{2}{c|}{ETTm1} & \multicolumn{2}{c|}{Weather} & \multicolumn{2}{c}{ECL} \\
      \cmidrule(lr){3-4}
      \cmidrule(lr){5-6}
      \cmidrule(lr){7-8}
      \cmidrule(lr){9-10}
      Var. &  & MSE & MAE & MSE & MAE & MSE & MAE & MSE & MAE \\
      \midrule
      Full &  & \textbf{0.429} & \textbf{0.452} & \textbf{0.413} & \textbf{0.412} & \textbf{0.310} & \textbf{0.326} & \textbf{0.198} & \textbf{0.282} \\
      $-$CA &  & 0.441 & 0.462 & 0.425 & 0.421 & 0.322 & 0.336 & 0.206 & 0.290 \\
      $-$G &  & 0.436 & 0.458 & 0.420 & 0.418 & 0.317 & 0.332 & 0.202 & 0.286 \\
      $-$Fus &  & 0.433 & 0.455 & 0.417 & 0.415 & 0.314 & 0.329 & 0.200 & 0.284 \\
      Coarse &  & 0.447 & 0.466 & 0.430 & 0.426 & 0.325 & 0.339 & 0.208 & 0.293 \\
      \bottomrule
    \end{tabular}
    \end{adjustbox}
    \endgroup
  \end{table}

Removing cross-attention ($-$CA) or fixing $\alpha{\equiv}1$ ($-$G) consistently raises both metrics, confirming that memory-conditioned refinement and learnable gating contribute beyond the coarse scaffold. Dropping concat--linear fusion ($-$Fus) hurts least yet remains positive on all datasets, showing that re-injecting $\mathbf{C}$ keeps layer-wise updates aligned with the global forecast scaffold. \emph{Coarse only}---skipping the cross-self stack entirely---incurs the largest error increase on every dataset, with the gap widest on high-dimensional ECL.

\subsubsection{Training objective.}
\label{sec:loss_ablation_ett_weather}

We fix the SACQ head and optimization protocol and vary only the loss paired with each readout. Table~\ref{tab:loss_objective} reports ETTh1, ETTm2, and Weather at $T_f\in\{96,720\}$. Keys: \textbf{F+slc} = flatten readout with scaled log-cosh; \textbf{S+MSE}, \textbf{S+lc}, and \textbf{S+slc} = SACQ with MSE, log-cosh, or scaled log-cosh, respectively. \textbf{Bold} marks the better MSE/MAE among the four recipes at each $(\mathrm{dataset},T_f)$.

\begin{table}[!t]
    \centering
    \caption{Training-objective ablation on ETTh1, ETTm2, and Weather ($T_f\in\{96,720\}$). PatchTST backbone; only the readout--loss pairing changes. \textbf{Bold}: better MSE/MAE among F+slc, S+MSE, S+lc, and S+slc at each setting.}
    \label{tab:loss_objective}
    \begingroup
    \fontsize{6}{7}\selectfont
    \setlength{\tabcolsep}{1.8pt}
    \renewcommand{\arraystretch}{0.92}
    \begin{adjustbox}{width=0.95\columnwidth,center}
    \begin{tabular}{@{}>{\raggedright\arraybackslash}p{3.8em}@{\hspace{-2pt}}r@{\hspace{1pt}}|cc|cc||cc|cc@{}}
      \toprule
       &  & \multicolumn{2}{c|}{F+slc} & \multicolumn{2}{c||}{S+MSE} & \multicolumn{2}{c|}{S+lc} & \multicolumn{2}{c}{S+slc} \\
      \cmidrule(lr){3-4}
      \cmidrule(lr){5-6}
      \cmidrule(lr){7-8}
      \cmidrule(lr){9-10}
      Dataset & $T_f$ & MSE & MAE & MSE & MAE & MSE & MAE & MSE & MAE \\
      \midrule
      \multirow{2}{*}{ETTh1} & 96 & 0.368 & 0.397 & {0.367} & {0.398} & 0.362 & 0.392 & \textbf{0.361} & \textbf{0.390} \\
       & 720 & 0.445 & 0.458 & {0.436} & {0.460} & 0.431 & 0.454 & \textbf{0.429} & \textbf{0.452} \\
      \midrule
      \multirow{2}{*}{ETTm2} & 96 & 0.165 & 0.253 & {0.161} & {0.253} & \textbf{0.159} & 0.245 & \textbf{0.159} & \textbf{0.244} \\
       & 720 & 0.361 & 0.383 & {0.361} & {0.381} & 0.359 & 0.377 & \textbf{0.358} & \textbf{0.375} \\
      \midrule
      \multirow{2}{*}{Weather} & 96 & 0.148 & 0.194 & {0.148} & {0.193} & 0.147 & 0.187 & \textbf{0.146} & \textbf{0.185} \\
       & 720 & 0.313 & 0.332 & {0.311} & {0.332} & 0.310 & 0.329 & \textbf{0.308} & \textbf{0.324} \\
      \bottomrule
    \end{tabular}
    \end{adjustbox}
    \endgroup
  \end{table}

Scaled log-cosh under SACQ (\textbf{S+slc}) is strongest on Weather and on MAE at long horizons; plain log-cosh (\textbf{S+lc}) can edge out on MSE in some ETT cells. Both log-cosh variants improve on flatten+scaled log-cosh and on SACQ+MSE. The robust loss and structured readout are complementary rather than interchangeable.

\subsubsection{Hyperparameter sensitivity.}
\label{sec:param_sensitivity}

We sweep the SACQ decoder depth $L\in\{1,2,3\}$ and the scaled log-cosh multiplier $k\in\{0.5,1.0,2.0,4.0\}$ in $\beta=k\,m$ (Eqs.~\eqref{eq:beta_median}--\eqref{eq:pred_loss_stable}), holding the PatchTST backbone and all other training settings fixed. Table~\ref{tab:param_sensitivity} reports clean-test MSE and MAE on ETTh1 and Weather at $T_f{=}720$; \textbf{bold} marks the default $(L,k)=(2,0.5)$.

\begin{table}[!t]
  \centering
  \caption{Hyperparameter sensitivity on clean test at $T_f{=}720$. \textbf{Bold}: default decoder depth $L{=}2$ and loss scale $k{=}0.5$ in $\beta=k\,m$.}
  \label{tab:param_sensitivity}
  \begingroup
  \fontsize{6}{7}\selectfont
  \setlength{\tabcolsep}{1.6pt}
  \renewcommand{\arraystretch}{0.92}
  \begin{adjustbox}{width=0.87\columnwidth,center}
  \begin{tabular}{@{}>{\raggedright\arraybackslash}p{3.6em}@{\hspace{-2pt}}c@{\hspace{1pt}}|cc|cc|cc|cc@{}}
    \toprule
    Dataset & $L$ & \multicolumn{2}{c|}{$k{=}0.5$} & \multicolumn{2}{c|}{$k{=}1.0$} & \multicolumn{2}{c|}{$k{=}2.0$} & \multicolumn{2}{c}{$k{=}4.0$} \\
    \cmidrule(lr){3-4}\cmidrule(lr){5-6}\cmidrule(lr){7-8}\cmidrule(lr){9-10}
     &  & MSE & MAE & MSE & MAE & MSE & MAE & MSE & MAE \\
    \midrule
    \multirow{3}{*}{ETTh1} & 1 & 0.431 & 0.451 & 0.438 & 0.456 & 0.445 & 0.463 & 0.444 & 0.458 \\
     & 2 & \textbf{0.430} & \textbf{0.450} & 0.429 & 0.452 & 0.431 & 0.456 & 0.436 & 0.456 \\
     & 3 & 0.431 & 0.450 & 0.438 & 0.455 & 0.437 & 0.458 & 0.441 & 0.458 \\
    \midrule
    \multirow{3}{*}{Weather} & 1 & 0.307 & 0.325 & 0.308 & 0.325 & 0.309 & 0.327 & 0.310 & 0.326 \\
     & 2 & \textbf{0.308} & \textbf{0.324} & 0.308 & 0.327 & 0.309 & 0.328 & 0.312 & 0.326 \\
     & 3 & 0.307 & 0.325 & 0.308 & 0.326 & 0.307 & 0.327 & 0.315 & 0.325 \\
    \bottomrule
  \end{tabular}
  \end{adjustbox}
  \endgroup
\end{table}

At $k{=}0.5$, clean-test error is almost insensitive to decoder depth: ETTh1 MSE varies by only 0.001 across $L\in\{1,2,3\}$ and Weather MSE by at most 0.001, with MAE similarly stable. As $k$ increases, the spread across $L$ widens. We therefore set $k{=}0.5$ with $L{=}2$ as the default: accuracy remains competitive without dataset-specific tuning of decoder depth.

\subsection{Robustness Evaluation}
\label{sec:robustness_eval}

Flatten readout couples all future steps through shared parameters, and MSE can overweight extreme supervision errors; both effects intensify under corrupted inputs or labels. We therefore evaluate SACQ along three stress axes---inference-time input corruption, training-set defects, and tail-shifted supervision---and inspect gate dynamics as a mechanistic follow-up.

\subsubsection{Inference-time input corruptions.}
\label{sec:corruption_severity_table}
We adopt the TSRBench protocol of Kim et al.~\cite{kim2026local}. All models are trained on clean data; at inference, combined \emph{spike} and \emph{level-shift} corruptions are injected into test \emph{input} windows while forecast targets remain clean. Severity levels $0$--$5$ follow~\cite{kim2026local}; all models use lookback $N{=}512$. Table~\ref{tab:corruption_severity} reports test MSE/MAE averaged over $T_f\in\{96,192,336,720\}$, comparing PatchSACQ against PatchLGA~\cite{kim2026local} and PatchTST~\cite{nie2023patchtst}.

\begin{table*}[!t]
  \centering
  \scriptsize
  \setlength{\tabcolsep}{2.2pt}
  \renewcommand{\arraystretch}{1.05}
  \caption{Test MSE/MAE under TSRBench combined spike and level-shift corruptions~\cite{kim2026local}. Models are trained on clean data; at inference, corruptions are injected into test inputs while forecast targets remain clean. Severity levels 0--5 follow~\cite{kim2026local}; metrics are averaged over $T_f\in\{96,192,336,720\}$ with $N{=}512$. \textbf{Bold}: best MSE/MAE among PatchSACQ, PatchLGA, and PatchTST at each $(\mathrm{dataset},\mathrm{severity})$.}
  \label{tab:corruption_severity}
  \begin{adjustbox}{width=\textwidth,center}
  \begin{tabular}{@{}c|r|cc|cc|cc||cc|cc|cc||cc|cc|cc@{}}
    \toprule
    \corruptionheader{ETTh1}{ETTh2}{Electricity}
    \multirow{6}{*}{\rotatebox{90}{\strut Severity\strut}} & 0 &
      \textbf{0.401} & \textbf{0.421} & 0.424 & 0.441 & 0.413 & 0.434 &
      \textbf{0.326} & \textbf{0.378} & 0.334 & 0.381 & 0.331 & 0.381 &
      \textbf{0.158} & \textbf{0.246} & 0.161 & 0.255 & 0.159 & 0.253 \\
    & 1 &
      \textbf{0.403} & \textbf{0.422} & 0.429 & 0.445 & 0.415 & 0.436 &
      \textbf{0.325} & \textbf{0.378} & 0.339 & 0.388 & 0.331 & 0.383 &
      0.169 & \textbf{0.257} & \textbf{0.168} & 0.264 & 0.169 & 0.265 \\
    & 2 &
      \textbf{0.407} & \textbf{0.428} & 0.431 & 0.446 & 0.420 & 0.442 &
      \textbf{0.331} & \textbf{0.386} & 0.346 & 0.396 & 0.340 & 0.394 &
      \textbf{0.178} & \textbf{0.267} & 0.179 & 0.276 & 0.184 & 0.279 \\
    & 3 &
      \textbf{0.460} & \textbf{0.462} & 0.477 & 0.477 & 0.475 & 0.476 &
      \textbf{0.344} & \textbf{0.396} & 0.357 & 0.405 & 0.357 & 0.406 &
      \textbf{0.192} & \textbf{0.279} & 0.193 & 0.288 & 0.201 & 0.293 \\
    & 4 &
      0.544 & \textbf{0.508} & \textbf{0.535} & 0.510 & 0.559 & 0.523 &
      \textbf{0.365} & \textbf{0.412} & 0.379 & 0.419 & 0.386 & 0.427 &
      \textbf{0.212} & \textbf{0.301} & 0.226 & 0.317 & 0.232 & 0.321 \\
    & 5 &
      \textbf{0.724} & \textbf{0.585} & 0.771 & 0.610 & 0.736 & 0.598 &
      \textbf{0.392} & \textbf{0.429} & 0.418 & 0.442 & 0.424 & 0.447 &
      \textbf{0.246} & \textbf{0.328} & 0.284 & 0.352 & 0.290 & 0.358 \\
    \midrule
    \corruptionheader{ETTm1}{ETTm2}{Weather}
    \multirow{6}{*}{\rotatebox{90}{\strut Severity\strut}} & 0 &
      \textbf{0.346} & \textbf{0.371} & 0.351 & 0.379 & 0.353 & 0.382 &
      \textbf{0.251} & \textbf{0.305} & 0.257 & 0.316 & 0.256 & 0.317 &
      \textbf{0.223} & \textbf{0.253} & 0.227 & 0.265 & 0.225 & 0.264 \\
    & 1 &
      \textbf{0.353} & \textbf{0.378} & 0.359 & 0.385 & 0.359 & 0.389 &
      \textbf{0.257} & \textbf{0.314} & 0.264 & 0.326 & 0.264 & 0.328 &
      \textbf{0.232} & \textbf{0.264} & 0.238 & 0.284 & 0.239 & 0.284 \\
    & 2 &
      \textbf{0.369} & \textbf{0.390} & 0.372 & 0.397 & 0.375 & 0.402 &
      \textbf{0.263} & \textbf{0.320} & 0.276 & 0.335 & 0.277 & 0.338 &
      \textbf{0.245} & \textbf{0.278} & 0.264 & 0.313 & 0.266 & 0.314 \\
    & 3 &
      0.528 & \textbf{0.467} & \textbf{0.519} & 0.468 & 0.614 & 0.507 &
      \textbf{0.284} & \textbf{0.336} & 0.304 & 0.355 & 0.308 & 0.360 &
      \textbf{0.256} & \textbf{0.288} & 0.301 & 0.338 & 0.306 & 0.339 \\
    & 4 &
      0.624 & \textbf{0.522} & \textbf{0.617} & 0.526 & 0.695 & 0.558 &
      \textbf{0.319} & \textbf{0.367} & 0.356 & 0.395 & 0.361 & 0.400 &
      \textbf{0.271} & \textbf{0.305} & 0.361 & 0.389 & 0.369 & 0.384 \\
    & 5 &
      \textbf{0.733} & \textbf{0.569} & 0.734 & 0.577 & 0.839 & 0.613 &
      \textbf{0.367} & \textbf{0.392} & 0.421 & 0.428 & 0.431 & 0.434 &
      \textbf{0.287} & \textbf{0.319} & 0.454 & 0.423 & 0.491 & 0.423 \\
    \bottomrule
  \end{tabular}
  \end{adjustbox}
\end{table*}

PatchSACQ retains the most best or tied-best cells as severity rises, while PatchTST degrades fastest on nearly every dataset; the margin widens markedly at severity~$\ge 3$. ETTm1 is the main exception, where PatchLGA occasionally achieves lower MSE at mid-to-high severity yet PatchSACQ still holds the best MAE.

\subsubsection{Training-set corruptions.}
\label{sec:robustness_corruption}
\begin{figure}[!t]
  \centering
  \includegraphics[width=\linewidth]{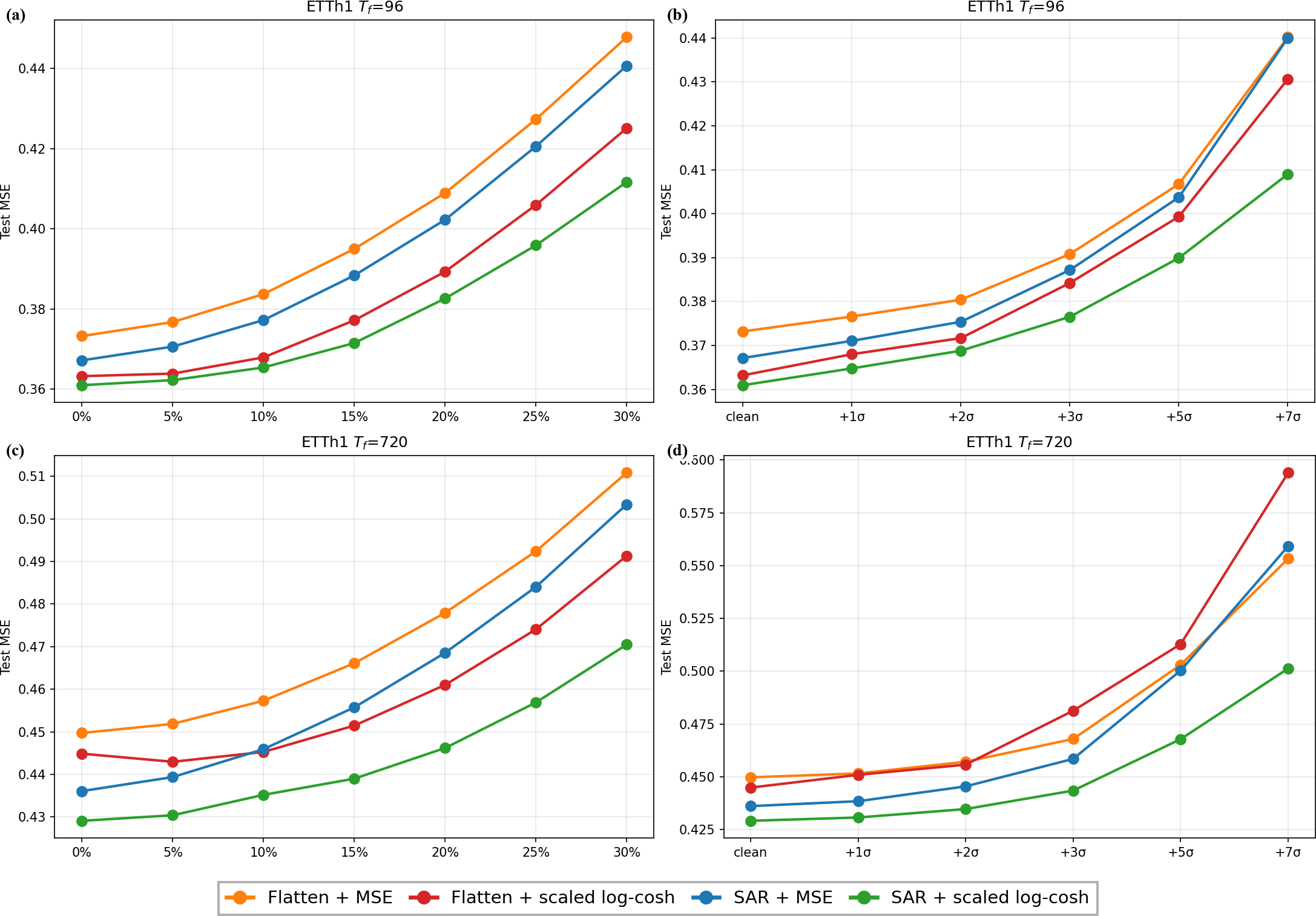}
  \caption{Robustness on ETTh1 under training-set corruption. \textbf{(a)} $T_f{=}96$, input masking: test MSE versus missing rate $\rho$ on historical windows. \textbf{(b)} $T_f{=}96$, target spikes: test MSE versus spike intensity (clean to +7$\sigma_y$). \textbf{(c)}--\textbf{(d)} Same sweeps at long horizon $T_f{=}720$. Curves compare Flatten+MSE, Flatten+scaled log-cosh, SAR+MSE, and SAR+scaled log-cosh; evaluation uses the clean test split.}
  \label{fig:corruption_etth1}
\end{figure}

As a complementary stress test, we inject custom defects into the \emph{training} split only; validation and test remain clean. \textbf{Input masking} zeroes a fraction $\rho\in\{0\%,5\%,\ldots,30\%\}$ of timesteps in each lookback window. \textbf{Target spikes} replace selected supervision steps with $y\leftarrow y+k\sigma_y$ ($k\in\{0,1,2,3,5,7\}$). Each corrupted model is retrained from scratch. Fig.~\ref{fig:corruption_etth1} on ETTh1 compares Flatten+MSE, Flatten+scaled log-cosh, SAR+MSE, and SAR+scaled log-cosh at $T_f\in\{96,720\}$.

All methods worsen with $\rho$, but structured readouts degrade more slowly; SAR+scaled log-cosh stays lowest, with SAR+MSE separating once $\rho\ge 15\%$. Under target spikes, SAR+scaled log-cosh has the shallowest slope from +3$\sigma_y$; at +7$\sigma_y$ the spread over flatten baselines is largest. At $T_f{=}720$, the same ordering holds despite higher absolute MSE.

\subsubsection{Causal bias under tail-shifted supervision.}
\begin{figure}[!t]
  \centering
  \includegraphics[width=\linewidth]{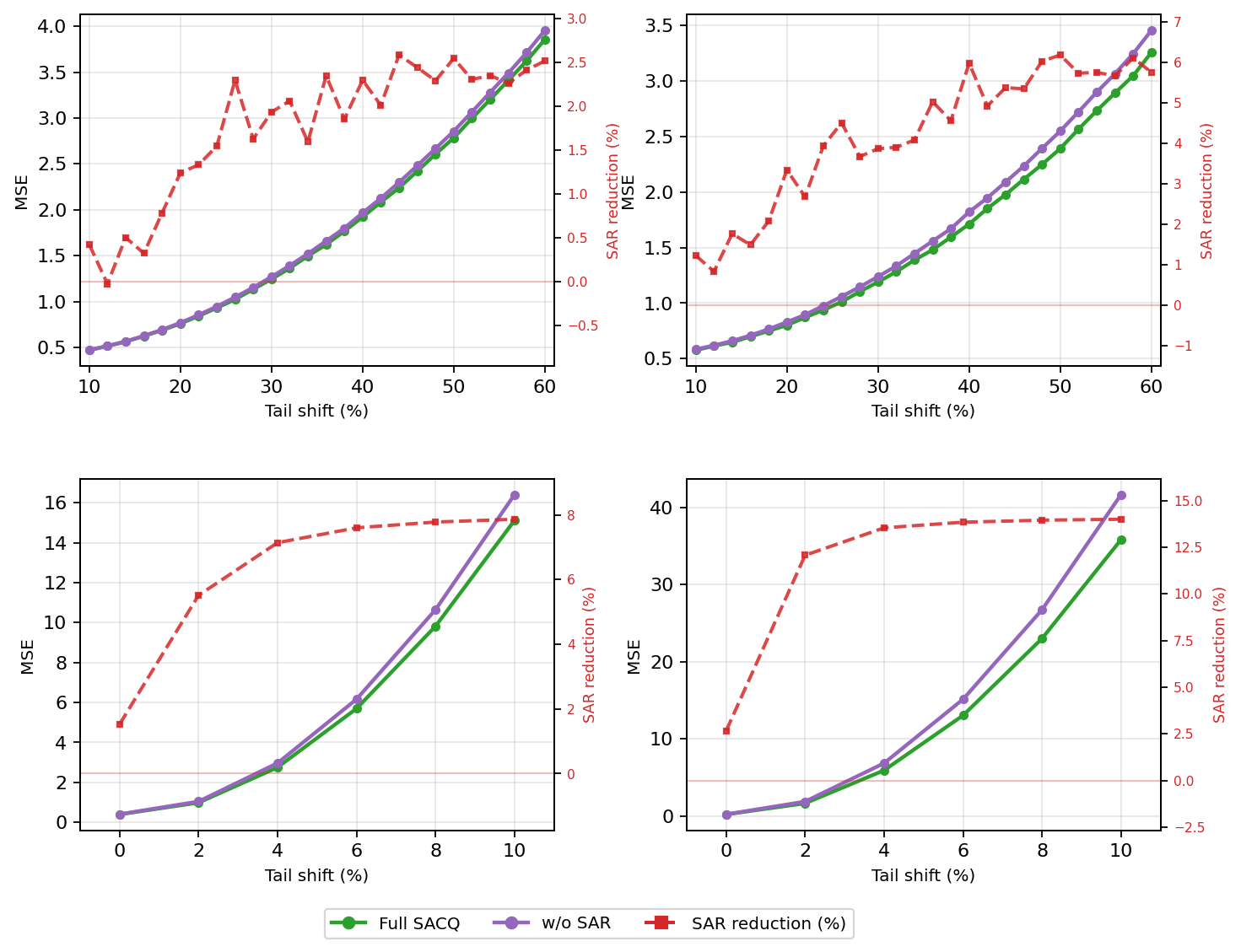}
  \caption{SAR ablation under training-set tail shift. \textbf{(a)--(b)} ETTh1 at $T_f{=}96$ and $720$. \textbf{(c)--(d)} Weather at $T_f{=}96$ and $720$. Solid curves: full SACQ (with SAR) and \emph{w/o SAR}; dashed curve: relative test-MSE reduction from SAR. Tail-shift rate increases the fraction of spike-corrupted supervision in the forecast tail during training only.}
  \label{fig:sar_tail_shift}
\end{figure}

To isolate SAR (Sec.~\ref{sec:sar}), we compare full SACQ with a variant that restores bidirectional self-attention among future patches (\emph{w/o SAR}). Both models use scaled log-cosh under a \emph{tail shift} that progressively replaces target steps in the forecast tail with spike-corrupted supervision during training (validation and test remain clean). Fig.~\ref{fig:sar_tail_shift} reports test MSE on ETTh1 and Weather at $T_f\in\{96,720\}$.

Full SACQ stays below \emph{w/o SAR} as tail shift intensifies, and the SAR gain grows with shift severity, especially at $T_f{=}720$ on ETTh1. These trends support the near-to-far inductive bias in Eq.~\eqref{eq:sar_mask}: restricting future-query interactions to strictly earlier patches limits contamination from later corrupted supervision steps.

\subsubsection{Gate dynamics as adaptive correction.}
\begin{figure}[!t]
  \centering
  \includegraphics[width=\linewidth]{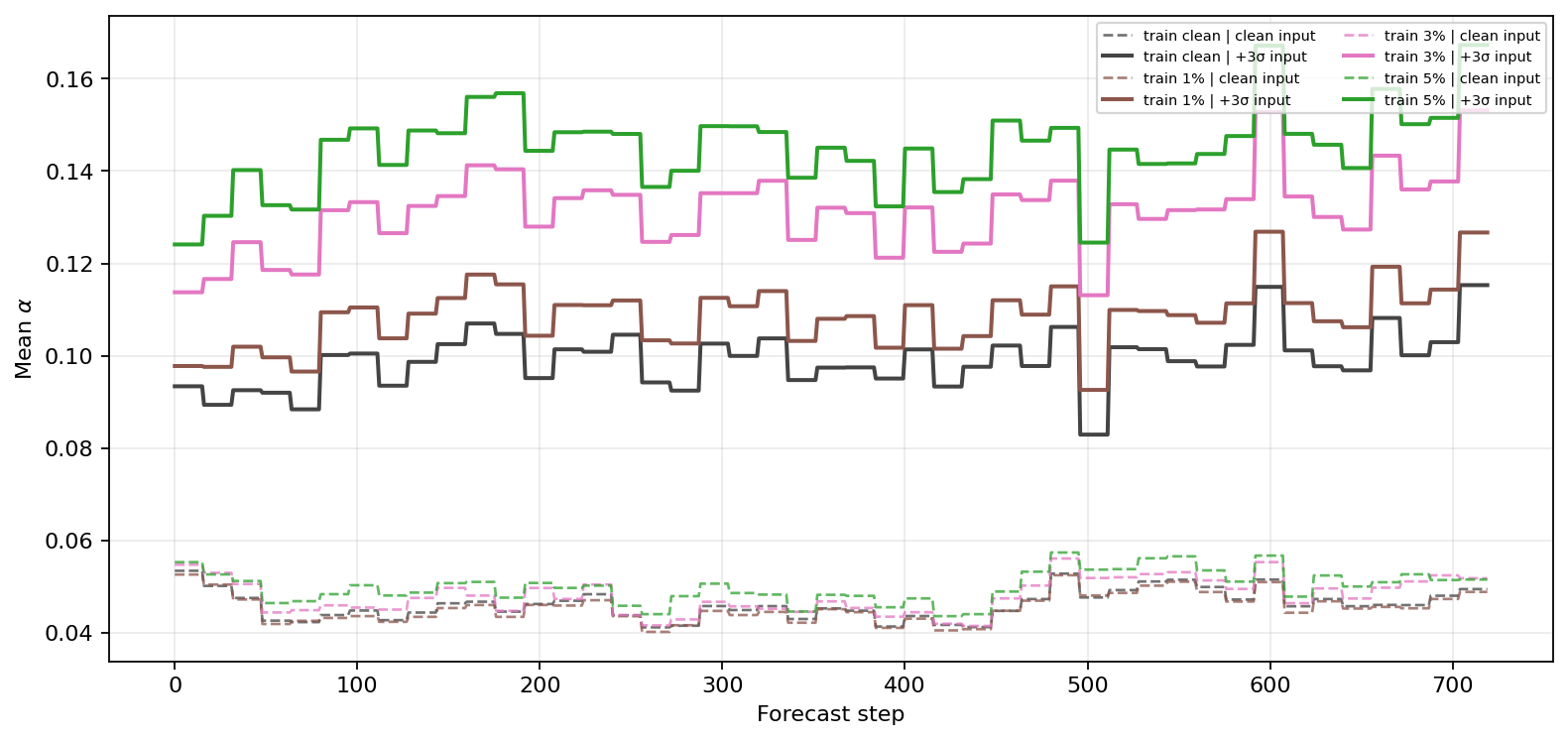}
  \caption{Horizon-wise mean gate $\alpha$ under clean versus spiked \emph{input} histories ($+3\sigma$ on the lookback window). \textit{Dashed:} clean test input. \textit{Solid:} $+3\sigma$ spiked test input. Colors index the training-set target-spike ratio ($0\%,1\%,3\%,5\%$).}
  \label{fig:alpha_horizon_spiked}
\end{figure}

To understand why SACQ tolerates corrupted inputs more gracefully (Table~\ref{tab:corruption_severity}), we track the horizon-wise mean of $\alpha_i$ in Eq.~\eqref{eq:residual_readout} on the clean test split while injecting $+3\sigma$ outlier spikes into test input histories. Models are trained with target-spike ratios of $0\%,1\%,3\%,5\%$ in the training split (validation/test otherwise clean). Fig.~\ref{fig:alpha_horizon_spiked} plots mean $\alpha$ across forecast steps.

Spiked input histories raise mean $\alpha$ above clean-input levels, and the gap widens with the training spike ratio; when test inputs remain clean, the $\alpha$ curves for different training ratios nearly overlap. Isolated spikes break local historical--future consistency in the coarse branch, so the gate increases $\alpha$ to allocate more gain to the attention-derived correction---consistent with the slower degradation under input masking in Fig.~\ref{fig:corruption_etth1}(a).

\subsection{Context Sensitivity}
\label{sec:context_sensitivity}

\subsubsection{Lookback length.}
\label{sec:lookback_sensitivity}

\begin{figure}[!t]
  \centering
  \includegraphics[width=\linewidth]{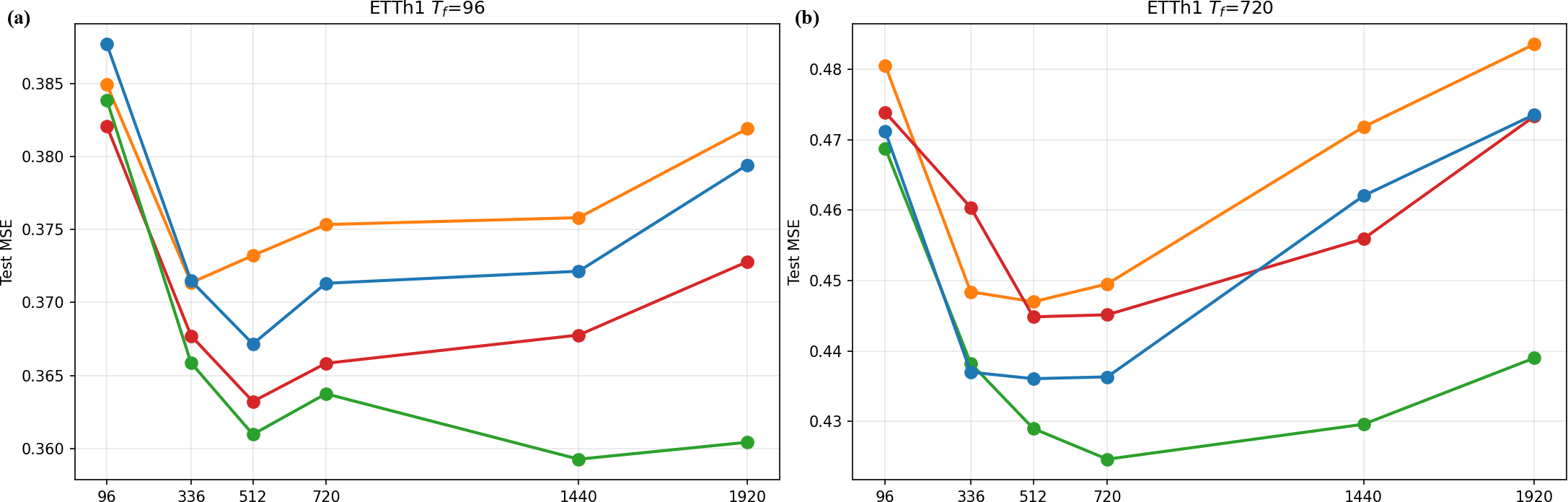}
  \caption{Lookback sensitivity on ETTh1. Test MSE versus input length $N$ for DLinear, PatchTST (flatten), SACQ+MSE, and SACQ+scaled log-cosh. \textbf{(a)} Short-horizon forecasting ($T_f{=}96$): all methods improve when $N$ grows from 96 to 336--512, then plateau. \textbf{(b)} Long-horizon forecasting ($T_f{=}720$): SACQ+scaled log-cosh remains stable at $N{=}1440$, whereas flatten baselines spike at the longest contexts.}
  \label{fig:lookback_sensitivity}
\end{figure}

Fig.~\ref{fig:lookback_sensitivity} varies lookback $N\in\{96,336,512,720,1440,1920\}$ on ETTh1 with fixed readout pairings per curve.

At $T_f{=}96$ [Fig.~\ref{fig:lookback_sensitivity}(a)], all configs gain history up to $N\in[512,720]$ then plateau; SACQ+scaled log-cosh is lowest for $N\ge 336$. At $T_f{=}720$ [(b)], flatten readouts spike near $N{=}1440$ while SACQ+scaled log-cosh stays stable, suggesting memory-conditioned decoding uses long contexts more effectively than a single global projection.

\subsection{Efficiency Analysis}
\label{sec:efficiency}
\begin{figure}[!t]
  \centering
  \includegraphics[width=\linewidth]{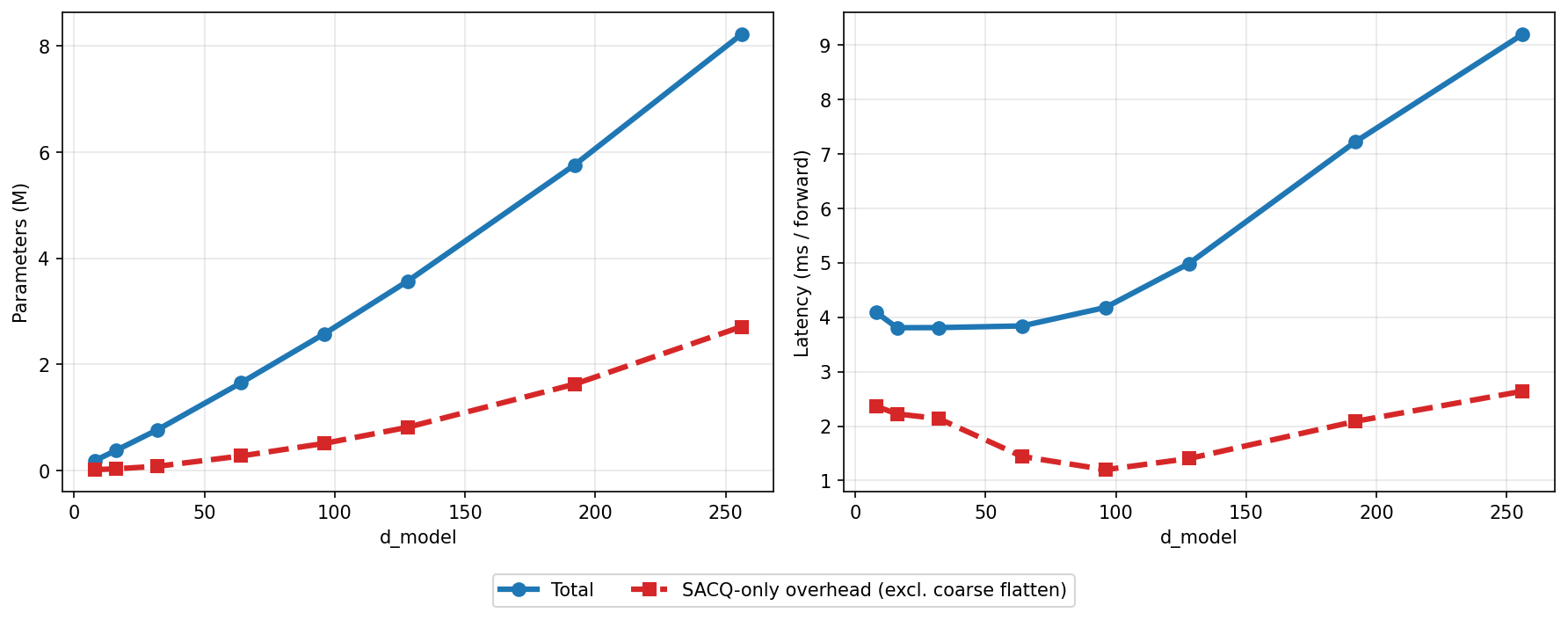}
  \caption{Scaling with backbone width $d_{\mathrm{model}}$. \textbf{Left:} parameter count (M). \textbf{Right:} forward-pass latency (ms). Blue solid: total model; red dashed: SACQ-only overhead excluding the coarse flatten map.}
  \label{fig:efficiency_dmodel}
\end{figure}

We profile how SACQ scales with backbone width by sweeping $d_{\mathrm{model}}\in\{8,16,32,64,96,128,192,256\}$ on a fixed PatchTST configuration ($T_f{=}96$, ETTh1 protocol). Fig.~\ref{fig:efficiency_dmodel} compares end-to-end totals against \emph{SACQ-only} overhead, defined as the decoder stack beyond the coarse flatten branch (cross--self attention, concat--linear fusion, and gated readout).

Both total parameters and end-to-end latency grow with $d_{\mathrm{model}}$, yet the SACQ-only curves rise more slowly: at $d_{\mathrm{model}}{=}256$, the incremental decoder adds roughly $2.7$M parameters and about $2.6$\,ms per forward pass, versus $\sim$8.2M parameters and $\sim$9.2\,ms for the full model. The extra cost is a width-dependent increment rather than an unfavorable super-linear penalty, supporting SACQ as a plug-in head with bounded deployment overhead.

\section{Conclusion and Future Work}
In this paper, we proposed SACQ, a plug-in head that recasts long-horizon forecasting as an explicit prediction mapping problem while keeping patch encoders unchanged and replacing flatten readout with staged decoding, conditional future queries, memory-guided refinement, and gated fusion, trained with batch-adaptive scaled log-cosh for stable multi-step optimization. Under a unified benchmark protocol, SACQ consistently improves accuracy over strong baselines, with controlled swaps indicating that both the head design and objective remain effective across encoder families. A main limitation is the added inference cost from extra attention and gating, reflected in a stable constant-factor latency increase despite modest parameter growth. Future work will target lower-latency SACQ variants, broader evaluation under severe distribution shifts and irregular horizons, and tighter integration with lightweight or retrieval-augmented backbones.


\bibliographystyle{IEEEtran}
\bibliography{sacq-references}

\end{document}